\documentclass[conference]{IEEEtran}
\IEEEoverridecommandlockouts
\usepackage{cite}
\usepackage{amsmath,amssymb,amsfonts}
\usepackage{algorithmic}
\usepackage{graphicx}
\usepackage{textcomp}
\usepackage{xcolor}
\usepackage{graphicx}
\usepackage{amsmath}
\usepackage{algorithmic}
\usepackage{latexsym}
\usepackage{makecell}
\usepackage{multirow}
\usepackage{subfig}
\usepackage{dblfloatfix}
\usepackage{float}
\usepackage{colortbl}
\usepackage{tabularx}
\usepackage{longtable}

\usepackage{array}
\usepackage[utf8]{inputenc} % allow utf-8 input
\usepackage[T1]{fontenc}    % use 8-bit T1 fonts
\usepackage[hidelinks]{hyperref}       % hyperlinks
\usepackage{url}            % simple URL typesetting
\usepackage{booktabs}       % professional-quality tables
\usepackage{nicefrac}       % compact symbols for 1/2, etc.
\usepackage{microtype}      % microtypography
\usepackage{fancyhdr}
\usepackage{lipsum} % mock text
\def\BibTeX{{\rm B\kern-.05em{\sc i\kern-.025em b}\kern-.08em
    T\kern-.1667em\lower.7ex\hbox{E}\kern-.125emX}}
\begin{document}

\title{Alzheimer’s Disease Diagnosis Using a Multimodal Approach with 3D MRI and PET

}

\author{\IEEEauthorblockN{Loukas Ilias, Anthi-Maria Vozinaki, Christos Ntanos, and Dimitris Askounis}
\IEEEauthorblockA{\textit{DSS Lab, School of ECE, NTUA, 15780 Athens, Greece} \\
lilias@epu.ntua.gr}
}

\maketitle

\begin{abstract}
Alzheimer’s disease (AD) is an irreversible neurodegenerative disorder and a leading cause of death worldwide. Early diagnosis plays an important part especially at the Mild Cognitive Impairment stage, where timely intervention can help slow its progression before it advances to AD. Neuroimaging data, like Magnetic Resonance Imaging (MRI) and Positron Emission Tomography (PET) scans, can help detect brain changes early by providing structural and functional brain changes related to the disease. Yet, many multimodal models still fuse MRI and PET with static concatenation and apply identical computation to all subjects, which limits robustness to patient/site heterogeneity and can waste computation. To address these limitations, we present the first study of combining 3D convolutional feature extractors with three fusion strategies - concatenation, Gated Multimodal Unit (GMU), and gated self-attention - and a sparsely gated Mixture-of-Experts (MoE) classifier that performs input-adaptive routing, activating only the most informative experts per case. Finally, we utilize Grad-CAM to visualize disease-related regions, ensuring model interpretability. Experiments are performed across three binary classification tasks (NC vs. MCI, MCI vs. AD, and NC vs. AD). Results show that GMU achieves accuracies of 80.46\% (NC vs. MCI) and 95.47\% (NC vs. AD), while gated self-attention attains 82.08\% on MCI vs. AD. Ablations show that removing the MoE consistently degrades accuracy across all tasks. These findings underscore the value of input-adaptive, multimodal modeling for AD diagnosis by leveraging the complementary nature of MRI and PET.
\end{abstract}


\section{Introduction}
Alzheimer’s disease (AD) is a progressive neurodegenerative disorder and the leading cause of dementia, accounting for 60–80\% of cases. It is characterized by memory loss, cognitive decline, and behavioral changes. In 2025\footnote{https://www.alz.org/alzheimers-dementia/facts-figures}, about 7.2 million Americans aged 65 and older are living with Alzheimer's, with numbers expected to rise in the future. At the molecular level, AD is marked by the buildup of beta-amyloid plaques outside neurons and abnormal tau protein tangles inside them. These changes impair synaptic function, leading to widespread brain cell death and shrinkage, especially in areas associated with memory, such as the hippocampus and cortex. Normal brain function is further compromised by the decreased ability of the brain to metabolize glucose, its main fuel \cite{mosconi2005brain}. As neuronal damage progresses, cognitive effects become more noticeable, with Mild Cognitive Impairment (MCI) marking the first recognizable stage where decline exceeds normal aging but has not yet significantly disrupted daily life \cite{KhanPetersen2024}. Individuals with MCI often experience memory lapses, difficulties with planning, or decision-making challenges. As MCI can be a precursor to AD, early detection is key. Recent advances in neuroimaging are making that possible with great precision. Among these, Positron Emission Tomography (PET) stands out for its ability to detect beta-amyloid plaques and can also measure glucose metabolism in the brain \cite{chapleau2022role}. Magnetic Resonance Imaging (MRI) complements PET by identifying structural changes, such as atrophy or shrinkage in critical areas, which are often affected in the early stages of the disease \cite{mcevoy2010quantitative}. Therefore, integrating both PET and MRI into routine clinical workflows provides both structural and pathological insights, enabling earlier interventions and more personalized treatment strategies \cite{shepherd2025clinical}. 

Existing studies come with limitations. Firstly, they have not adequately studied multimodal approaches that combine MRI and PET modalities. Even in cases where they have been applied, the integration of different modalities is usually limited to a simple concatenation, early or late fusion strategies \cite{venugopalan2021multimodal}, overlooking the interactions between them. Moreover, a significant limitation is the lack of dynamic adaptability to input data. Most approaches rely on dense layers for the final classification of subjects \cite{10478896}, which, due to their static nature, struggle to adapt to more complex and heterogeneous data. This is a critical issue in the case of MCI, a stage that is inherently associated with a high diagnostic complexity. Another key limitation involves the explainability of the models. Most developed models function as "black boxes," where they receive input data and produce output results without providing information about the features that contributed to their final decision \cite{9490307,10822255}. This poses a significant problem in clinical practice, as for a doctor to evaluate the model’s results, it is essential to know which brain regions are considered pathological. The lack of transparency undermines trust and complicates the implementation of these methods in diagnosis and clinical decision-making. 

In order to tackle these limitations, we present a study that uses both MRI and PET images, multimodal fusion methods and a MoE model into a single end-to-end trainable deep neural network. Firstly, we apply a series of preprocessing steps (e.g., skull stripping, registration to template). Then, we extract features from both imaging modalities using a 3D CNN. Next, we apply three different fusion techniques (Concatenation, Gated Multimodal Unit \cite{10.1007/s00521-019-04559-1}, Gated Self-Attention \cite{yu2019multimodal}) to capture both inter- and intra-modal interactions. In order to improve classification performance and computational efficiency, we integrate a Mixture of Experts \cite{shazeer2017} model, which dynamically selects the most relevant subnetworks for each prediction. Finally, we use Grad-CAM \cite{Selvaraju_2017_ICCV} to visualize which regions of the brain influence the model’s final decision. Results show multiple advantages of our proposed approach over state-of-the-art ones.

Our main contributions can be summarized as follows:
\begin{itemize}
    \item We present the first study utilizing MRI and PET images and incorporating fusion methods and a sparsely gated MoE layer in a single end-to-end deep neural network.
    \item We use three different fusion approaches and compare their performances.
    \item We apply Grad-CAM to visualize and interpret the model’s decision-making process.
    \item We conduct ablation experiments to explore the effectiveness of our introduced approach.
\end{itemize}

\section{Related Work} \label{related_work_section}

\subsection{Unimodal Approaches}
The study by \cite{lebedev2014random} investigated the use of Random Forest (RF) classifiers for detecting AD and distinguishing it from healthy controls using structural MRI data. Recursive feature elimination was employed to optimize feature selection, and models were trained using RF with different morphometric modalities. The model with the best performance achieved a sensitivity of 88.6\%, specificity of 92\%, and an area under the ROC curve of 0.94 for distinguishing AD from Normal Cognition (NC) in the ADNI dataset. 

The study by \cite{payan2015predicting} applied 3D CNNs to the classification of AD using neuroimaging data. The methodology involved preprocessing MRI scans from the ADNI dataset to normalize voxel intensities and reduce intersubject variability. Sparse autoencoders were first trained to learn feature representations from 3D patches of brain images, which were then used to initialize the 3D-CNN. The CNN architecture consisted of convolutional, pooling, and fully connected layers, allowing the model to capture local 3D patterns and hierarchical features directly from the volumetric MRI data. Their model achieved an accuracy of 95.39\% for AD vs. NC, 86.84\% for AD vs. MCI, and 92.11\% for MCI vs. NC, outperforming the 2D-CNN approach and many traditional machine learning methods. 

The study by \cite{sarraf2016classification} employed 2D CNNs, specifically the LeNet-5 architecture, to classify AD using functional MRI (fMRI) data. The data, obtained from the ADNI dataset, were preprocessed through standard pipelines, including motion correction, skull stripping, and spatial smoothing, before being converted into 2D JPEG images. These images were then labeled for binary classification (AD vs. NC) and processed through the CNN. The network was trained using 60\% of the data, validated on 20\%, and tested on the remaining 20\%. The model achieved a mean classification accuracy of 96.86\% across five runs. 

The authors in \cite{zhou2025deep} utilized 3D MRI data of the ADNI database and designed a deep neural network consisting of a 3D CNN followed by a fine-tuned Video Swin Transformer. An ensemble approach consisting of CNN-based classifiers was exploited in \cite{fathi2024deep}. The authors used MRI as input. A hybrid approach was introduced in \cite{matlani2024bilstm}. Specifically, the authors used MRI data, applied feature extraction and selection techniques, and then trained a hybrid BiLSTM with Artificial Neural Network (ANN). Similarly, the authors in \cite{9098621} used MRI and passed this data through a deep neural network consisting of CNNs and convolutional long short-term memory (CLSTM).

\subsection{Multimodal Approaches}

The study by \cite{liu2014early} presents a deep learning-based framework for diagnosing AD and MCI, using multimodal neuroimaging data, specifically MRI and PET, from the ADNI database. The proposed method utilizes stacked sparse auto-encoders for dimensionality reduction and data fusion, combined with a softmax regression layer for multi-class classification. The model achieved an accuracy of 87.76\%, sensitivity of 88.57\%, and specificity of 87.22\% for binary classification of AD vs NC. Additionally, in the 4-class classification task (NC, MCI converters, MCI non-converters, and AD), the framework attained an accuracy of 47.42\%. 

Ref. \cite{lu2018mmdnn} proposed a deep learning framework, the Multimodal and Multiscale Deep Neural Network (MMDNN), for the early diagnosis of AD. Using neuroimaging data (T1-MRI and FDG-PET scans) from the ADNI, the framework combines this data at multiple scales. The method involves preprocessing the images to extract multiscale patch-wise features and training independent neural networks for each scale and modality, followed by a feature-fusion network to generate predictions. For the classification of NC vs AD, the MMDNN achieved 84.6\% accuracy, with a sensitivity of 80.2\% and specificity of 91.8\%. Additionally, the framework also showed high performance in predicting conversion from MCI to AD within 1-3 years prior to diagnosis. 

The study by \cite{huang2019multimodal} leveraged also multimodal imaging data, specifically T1-weighted MRI and FDG-PET. The methodology centered around the extraction of 3D patches from the hippocampal region, a key area of atrophy in AD, allowing the model to capture complementary structural and metabolic features. The study utilized the ADNI dataset, comprising 731 NC subjects, 647 AD patients, 441 stable Mild Cognitive Impairment (sMCI) subjects, and 326 progressive MCI (pMCI) subjects. The model achieved classification accuracies of 90.10\% for AD vs. NC, 87.46\% for NC vs. pMCI, and 76.90\% for sMCI vs. pMCI.

The authors in \cite{song2021imagefusion} proposed a multimodal image fusion method to enhance the diagnosis of AD by creating a new composite imaging modality called GM-PET, which combines structural information from MRI and metabolic information from FDG-PET, with a focus on the gray matter (GM) region critical for AD diagnosis. Using the ADNI dataset, the authors pre-processed the data through skull stripping, registration, and segmentation to isolate the GM region. This GM region was then used to fuse complementary information from MRI and PET into the GM-PET modality, which retains brain structure and metabolic data while eliminating noise. Two classification networks, a 3D Simple CNN and a 3D Multi-Scale CNN, were employed to evaluate the fused modality. The GM-PET modality demonstrated significant improvements in diagnostic accuracy, achieving 94.11\% accuracy, 93.33\% sensitivity, and 94.27\% specificity for AD vs. NC classification, and 88.48\% accuracy for MCI vs. NC classification. 

The authors in \cite{golovanevsky2022multimodal} proposed the Multimodal Alzheimer’s Disease Diagnosis framework (MADDi), a deep learning-based system for diagnosing AD and MCI using imaging (MRI), genetic (SNPs), and clinical data from the ADNI dataset. The model utilized modality-specific neural network backbones (a CNN for MRI and fully connected networks for clinical and genetic data), followed by self-attention layers to identify critical intra-modality features and cross-modal attention layers to capture interactions between modalities. MADDi achieved an accuracy of 96.88\%, with an F1-score of 91.41\%. Unimodal models achieved lower accuracies (clinical: 80.59\%, genetic: 77.78\%, imaging: 92.28\%), highlighting the advantage of multimodal fusion. 

Ref. \cite{castellano2024automated} conducted a study on automated AD detection using a multimodal approach that integrates 3D MRI and amyloid PET imaging, along with transfer learning strategies. Using the OASIS-3 dataset, the researchers developed CNN models to evaluate unimodal (MRI or PET) and multimodal configurations. Transfer learning was implemented by pretraining on 3D PET data and fine-tuning on MRI data, and vice versa, but these methods did not surpass uni-modal or multimodal models, suggesting limited cross-modality feature applicability. The best performing model, a multimodal fusion model, achieved an accuracy of 95.00\%, sensitivity of 93.33\%, and specificity of 96.66\%, outperforming unimodal methods. MRI scans alone demonstrated stronger diagnostic performance than PET, but the integration of MRI and PET captured complementary features, significantly enhancing diagnostic accuracy. Grad-CAM explainability analysis identified critical brain regions associated with AD, such as the medial temporal lobe and frontal gyrus, emphasizing the clinical relevance of the findings. 

PET and MRI data were used in \cite{10478896}. Firstly, the authors employed depth-wise separable convolution layers to extract features from the two modalities. Next, middle fusion was implemented using a mix skip connection convolution block. Specifically, after the features of one modality are extracted, the features of the other modality are concatenated using skip connections. Finally, a SW-Conv block (two depthwise separable convolutional layers) followed by fully connected layers was used. A multimodal approach based on cross-attention mechanism was introduced in \cite{10822255}. Specifically, after extracting MRI and PET features through CNN layers, the authors used a cross-attention layer. A different approach was proposed by \cite{9490307}. Firstly, the authors used Generative Adversarial Neural Networks (GANs) to generate the missing modality (PET). Next, they trained a deep neural network with pathwise transfer blocks combining the features of PET and MRI data.

\section{Dataset and Preprocessing}
The data utilized in this study are obtained from the Alzheimer’s Disease Neuroimaging Initiative (ADNI) dataset (https://adni.loni.usc.edu/). ADNI is a large-scale, longitudinal, multicenter study aimed at developing clinical, imaging, genetic, and biochemical biomarkers for the early diagnosis and monitoring of Alzheimer’s disease (AD). The dataset includes multimodal data collected and analyzed primarily from North American participants, covering various stages of the disease \cite{jack2008adni}.

In this study, subjects with both T1-weighted MRI (specifically MPRAGE scans for their superior quality) and FDG-PET scans are selected, ensuring the scans were acquired closely in time to improve alignment between the modalities. The final cohort consists of 379 subjects from the ADNI dataset: 73 with AD, 188 with MCI, and 118 NC individuals. 

The MRI and FDG-PET images in the ADNI dataset have already undergone several preprocessing steps. We exploit this preprocessed data in our study.

MRI data have been processed via the following steps: 
\begin{itemize}
    \item \textbf{Gradwarp:} Corrects image geometry distortion caused by the gradient model.
    \item \textbf{B1 non-uniformity correction:} Addresses intensity variations using B1 calibration scans.
    \item \textbf{N3 histogram peak-sharpening algorithm:} Reduces intensity non-uniformity.
\end{itemize}

The baseline FDG-PET scans are processed through the following steps:
\begin{itemize}
    \item \textbf{Co-registration of dynamic frames:} Six 5-minute FDG-PET frames acquired 30–60 minutes post-injection are co-registered to the first frame to reduce the effects of patient motion.
    \item \textbf{Averaging:} The co-registered frames are averaged.
    \item \textbf{Standardization of image and voxel size:} The averaged image is reoriented into a standard 160 × 160 × 96 voxel grid with 1.5 mm cubic voxels, corrected for anterior commissure–posterior commissure alignment, and intensity-normalized using a subject-specific mask so that the average voxel value within the mask equals one.
    \item \textbf{Uniform resolution:} The normalized image is smoothed using a scanner-specific filter to achieve a uniform isotropic resolution of 8 mm full width at half maximum. 
\end{itemize}

Next, similar to \cite{song2021imagefusion}, we apply some additional preprocessing steps as follows:
\begin{itemize}
    \item The preprocessing of MRI scans begins with skull-stripping, implemented using the FSL (FMRIB Software Library) package, specifically the Brain Extraction Tool (BET). Skull-stripping removes non-brain tissues such as the skull, scalp, and dura mater, isolating the brain structure for more focused analysis. In the processing pipeline, a threshold parameter controls the aggressiveness of tissue removal and is set to 0.5 for balanced extraction. Bias field correction is also applied to improve the clarity of the extracted brain region, enhancing the overall quality of the output. The skull-stripped MRI (SS-MRI) images are then affine transformed to the MNI152 space, a widely adopted universal brain atlas template. This transformation employs the FLIRT (FMRIB’s Linear Image Registration Tool) module within the FSL package, which applies a linear affine transformation to correct for spatial discrepancies between subjects. The registration process aligns MRI scans by correcting translations, rotations, and scaling, standardizing the images to the consistent orientation and voxel dimensions of MNI152 space. 
    \item For FDG-PET scans, preprocessing begins with PET skull-stripping to isolate the brain structure, followed by co-registration to their corresponding MNI-aligned MRI images. Both steps are analogous to those performed for MRI scans, ensuring that the PET images adopt the same spatial orientation and voxel resolution (e.g., $1.0 \times 1.0 \times 1.0$ mm) as their MRI counterparts. This alignment guarantees consistency across modalities, facilitating multimodal integration and enabling the model to effectively learn spatial relationships inherent in the data. To improve computational efficiency, the co-registered PET and MRI images are resized to a lower resolution of $160 \times 180 \times 80$. 
\end{itemize}

% Figure~\ref{fig:process} illustrates the preprocessing pipeline for both MRI and PET data of the same subject, showcasing the steps of skull-stripping and registration to the MNI152 template, alongside the spatial alignment results in transverse, sagittal, and coronal planes for each modality. The displayed subject belongs to the NC group. 

% \begin{figure}[H]
%     \centering
%     \includegraphics[width=0.5\textwidth]{figures/preprocessing.png} 
%     \caption{Preproccesing pipeline for MRI and PET images} 
%     \label{fig:process} 
% \end{figure}

\section{Methodology}

\begin{figure}[H]
    \centering
    \includegraphics[width=1.0\columnwidth]{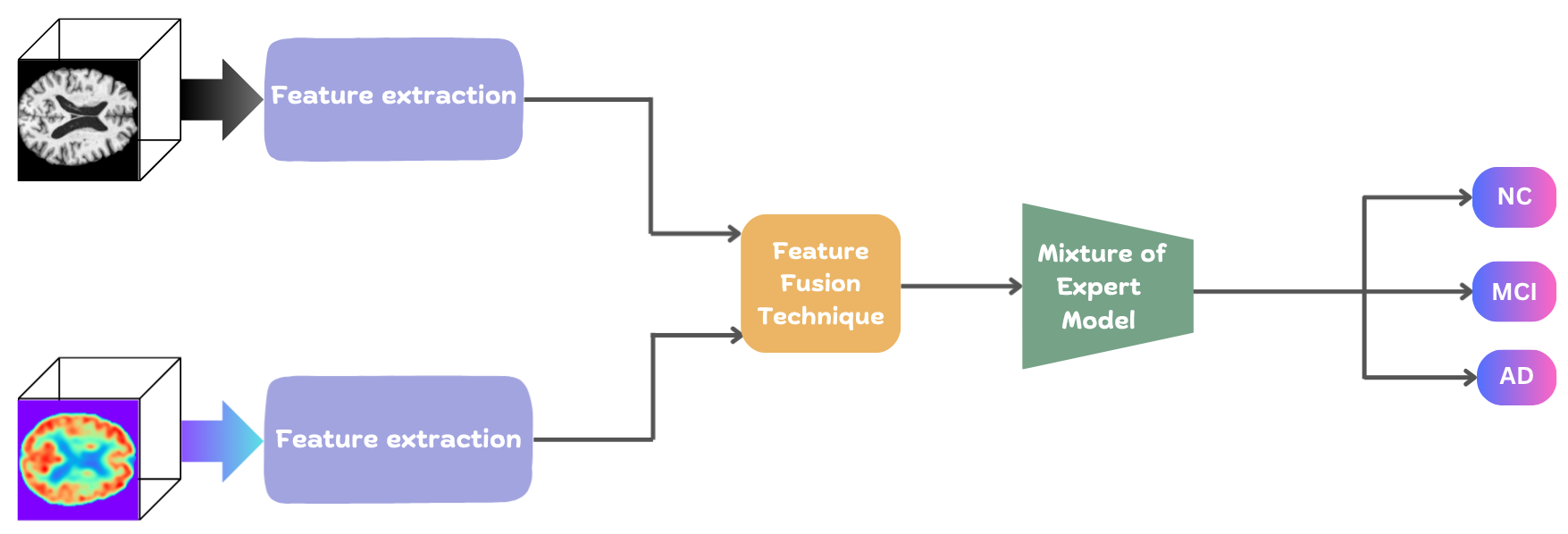} 
    \caption{Methodology pipeline} 
    \label{fig:methodology} 
\end{figure}

Our methodology is illustrated in Fig.~\ref{fig:methodology}.

\subsection{Feature Extraction - CNN Architecture}
\begin{figure}[H]
    \centering
    \includegraphics[width=1.0\columnwidth]{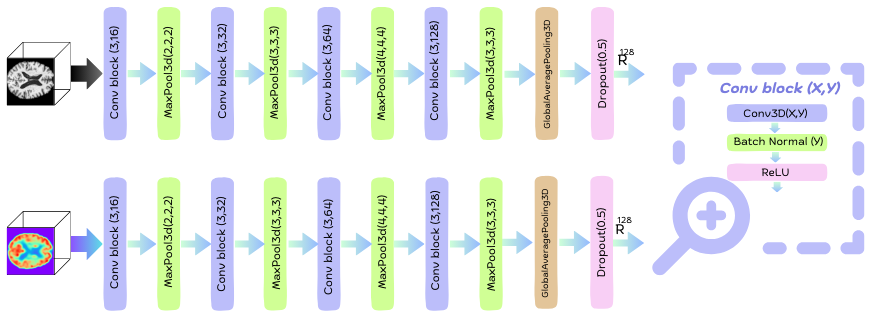} 
    \caption{The CNN architecture used in this study} 
    \label{fig:cnn3d} 
\end{figure}

As shown in Figure~\ref{fig:cnn3d}, the CNN architecture used in this study consists of four 3D convolutional layers, each with a kernel size of $3 \times 3 \times 3$. These layers are followed by batch normalization to stabilize training and ReLU activation to introduce non-linearity. The first convolutional layer generates eight feature maps, with subsequent layers producing 16, 32, 64 and 128 feature maps, respectively. Max-pooling layers are applied after each convolutional block to progressively reduce spatial dimensions while preserving key features. The pooling operations use kernel sizes of $2 \times 2 \times 2$, $3 \times 3 \times 3$, and $4 \times 4 \times 4$. In the final stage, a global average pooling layer further compresses the spatial dimensions, preparing the feature maps for the next processing steps. Dropout regularization is applied after pooling layers to reduce the risk of overfitting.
The architecture features two separate but identical pathways for MRI and PET images, each designed to extract distinct structural and functional information from the respective modalities. Both pathways culminate in 128 feature maps. 

\subsection{Fusion}
\begin{itemize}
\item \textbf{Concatenation}: It directly combines MRI and PET features into a single feature vector. This straightforward yet effective approach preserves all the information from both modalities, serving as a baseline for comparison with more advanced methods. The resulting concatenated vector has a dimensionality of 256, as each modality contributes 128 features. 

% \begin{figure}[H]
%     \centering
%     \includegraphics[width=0.3\textwidth]{figures/concat.png} 
%     \caption{Simple concatenation of the MRI and PET features} 
%     \label{fig:conc} 
% \end{figure}

\item \textbf{Gated Multimodal Unit (GMU) \cite{10.1007/s00521-019-04559-1}}. It learns to emphasize the more informative modality while suppressing the less relevant one, enabling effective and selective fusion of multimodal data. A gating mechanism generates weights to control the flow of information from each modality. The GMU takes 128 features from MRI and 128 features from PET as input and produce a unified output of 128 features. The equations governing the GMU framework are presented here: $h_m = \tanh(W_m f_m + b_1),
    h_p = \tanh(W_p f_p + b_2), 
    z = \sigma(W_z [f_m; f_p] + b_z), 
    h = z \odot h_m + (1 - z) \odot h_p$.

\noindent where:
\begin{itemize}
    \item \(f_p\) and \(f_m\) are the input features from the PET and MRI respectively. 
    \item \(W_p, W_m, W_z \in \mathbb{R}^{128}\) are learnable weight matrices for the respective modalities (PET and MRI) and gating mechanism respectively.
    \item \(z\) is the gating vector, computed via the sigmoid function (\(\sigma\)), which determines the relative importance of each modality.
    \item \(h_p\) and \(h_m\) are the transformed representations of the input features, passed through a \(\tanh\) activation function.
    \item \(h\) is the final fused representation, computed as a weighted combination of \(h_p\) and \(h_m\), with the gating vector \(z\) controlling the contribution of each modality.
\end{itemize}

% \begin{figure}[H]
%     \centering
%     \includegraphics[width=0.3\textwidth]{figures/gmu.png} 
%     \caption{GMU for the concatenation of MRI and PET features}  
%     \label{fig:gmu} 
% \end{figure}

\item \textbf{Gated Self-Attention}. This approach allows the model to capture both intra-modal and inter-modal interactions simultaneously while retaining only the most relevant connections. Given the two modalities, we extract features before applying global average pooling (see Fig.~\ref{fig:cnn3d}) and then we concatenate their representations:

\begin{equation}
    Z = [MRI; PET]
\end{equation}

% \begin{figure}[H]
%     \centering
%     \includegraphics[width=0.5\textwidth]{figures/nogap2.png} 
%     \caption{Feature extraction without Global Average Pooling} 
%     \label{fig:nogap} 
% \end{figure}

Similar to \cite{10.3389/fnagi.2022.830943}, this representation is then used for computing the query (\( Q \)), key (\( K \)), and value (\( V \)) matrices:

\begin{equation}
    Q = Z, \quad K = Z, \quad V = Z.
    \label{eq: att1}
\end{equation}

Next, we adopt the gating model introduced by \cite{yu2019multimodal} as follows:$M = \sigma \left (FC^g \left(FC^g _q \left(Q\right)\odot FC^g _k \left(K\right)\right)\right )$, where $FC^g _q, FC^g _k \in \mathbb{R}^{d \times d_g}$, $FC^g \in \mathbb{R}^{d_g \times 2}$ are three fully-connected layers, and $d_g$ denotes the dimensionality of the projected space. $\odot$ denotes the element-wise product function and $\sigma$ the sigmoid function. In addition, $M \in \mathbb{R}^{m \times 2}$ corresponds to the two masks $M_q \in \mathbb{R}^m$ and $M_k \in \mathbb{R}^m$ for the features $Q$ and $V$ respectively.

Next, the two masks $M$ and $K$ are tiled to $\Tilde{M_q}, \Tilde{M_k} \in \mathbb{R}^{m \times d}$ and then used for computing the attention map as following:
\begin{equation}
    A^g = softmax \left (\frac{\left(Q \odot \Tilde{M_q}\right)\left(K \odot \Tilde{M_k}\right)^T}{\sqrt{d}} \right )
    \label{eq_gating_model}
\end{equation}

\begin{equation}
    H = A^g V
    \label{a_g}
\end{equation}

Then, the output \textit{H} is passed through a global average pooling layer followed by a dense layer with 128 units and a ReLU activation function. 

\end{itemize}
% \begin{figure}[H]
%     \centering
%     \includegraphics[width=0.4\textwidth]{figures/attention.png} 
%     \caption{Attention mechanism after the concatenation of the MRI and PET features} 
%     \label{fig:att} 
% \end{figure}
\subsection{Mixture of Experts}
\begin{figure}[H]
    \centering
    \includegraphics[width=0.95\columnwidth]{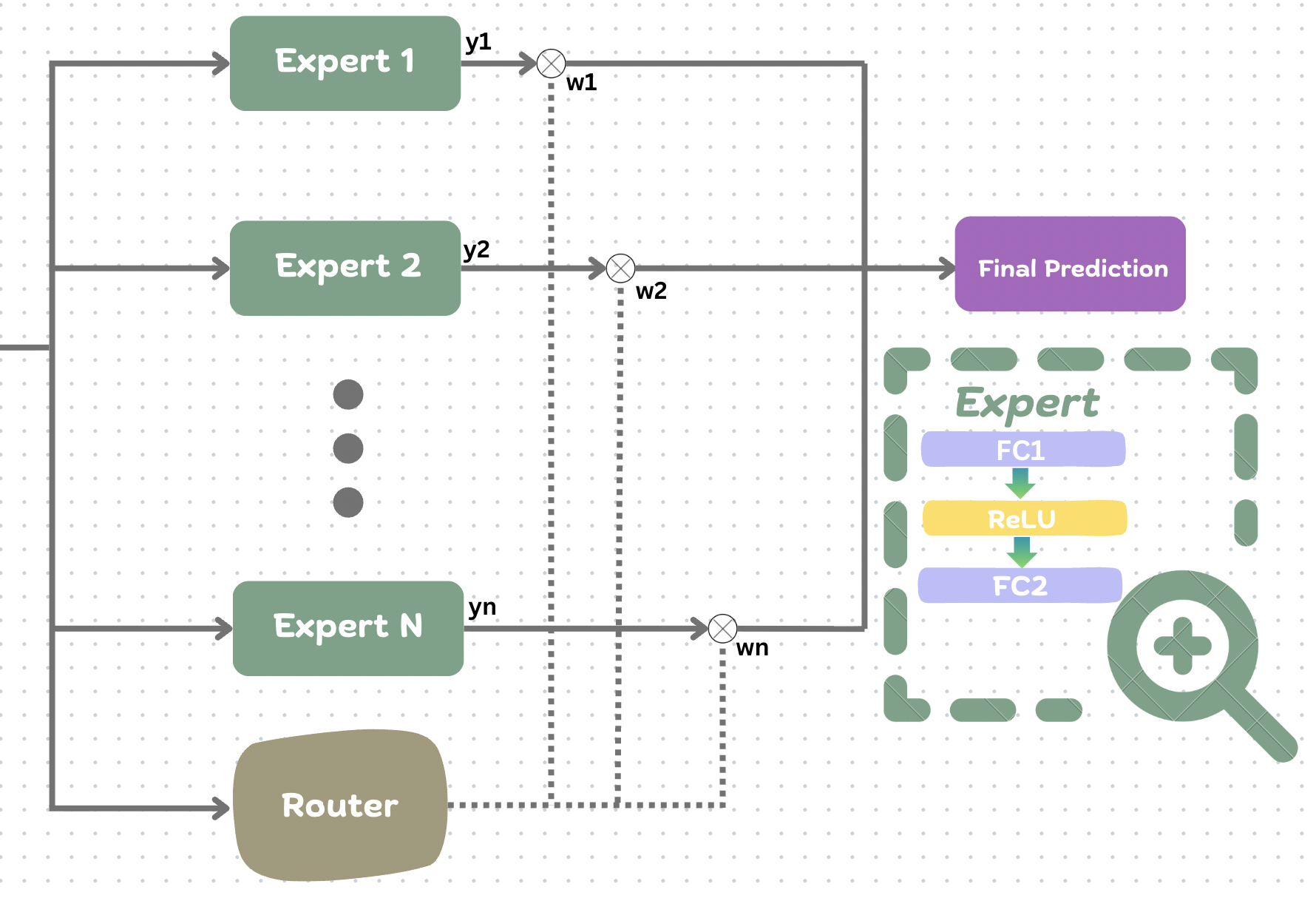} 
    \caption{MoE architecture of this study} 
    \label{fig:moe} 
\end{figure}

We use a sparsely gated MoE layer \cite{shazeer2017}. \( n \) experts process the extracted features from the three previously discussed fusion methods. As shown in Fig.~\ref{fig:moe}, each expert consists of a fully connected layer followed by a ReLU activation function, and a second fully connected layer for final classification. Next, we describe the method for obtaining the gating coefficients. 

We add noise to the input ($H(x)_i = (x \cdot W_g)_i + StandardNormal() \cdot Softplus \left(\left(x \cdot W_{noise} \right)_i \right)$), where $StandardNormal()$ indicates standard normal distribution. Next, we keep only the top-\textit{k} values (see Eq.~\ref{keep_topk}). Adding noise facilitates load balancing among the experts, while keeping only the top-\textit{k} most relevant experts saves computation, since only a few of the experts are activated via a gating network.

\begin{equation} 
        \resizebox{\linewidth}{!}{$
        \text{KeepTopK}(v, k)_i = \begin{cases} v_i & \text{if } v_i \text{ is in the top } k \text{ elements of } v, \\ 
            -\infty & \text{otherwise.}
            \end{cases}$}
            \label{keep_topk}
        \end{equation}

Finally, we apply the softmax activation function to get the coefficients  $G(x) = Softmax \left(KeepTopK\left(H\left(x\right),k\right)\right) $.

Next, we describe the \textbf{loss function}, which is minimized in this study. 
\begin{itemize}
    \item $L_{imp}$: This loss encourages the gate to assign similar weights to all experts, preventing a few from dominating. It is computed as the squared coefficient of variation of the experts’ importance values: $
            L_{\text{Imp}}(\mathcal{B}) = \text{CV} \left( \left\{ \text{Imp}_i(\mathcal{X}) \right\}_{i=1}^n \right)^2
            $, 
        where 
        $
            \text{CV}(\cdot) = \frac{\text{Std}(\cdot)}{\text{Mean}(\cdot)},$ and $
            \text{Imp}_i(\mathcal{X}) = \sum_{x \in \mathcal{X}} G(x)
            $.
\item         $L_{Load}$: This loss encourages balanced routing, so each expert processes the same number of training examples.
        $
            L_{\text{Load}}(\mathcal{B}) = \text{CV} \left( \left\{ \text{Load}_i(\mathcal{X}) \right\}_{i=1}^n \right)^2
        $, where $            \text{Load}_i(\mathcal{X}) = \sum_{x \in \mathcal{X}} P_i(x)$ . $Load_i$ denotes the number of training examples per expert and $P_i(x)$ denotes the probability that $G(x)_i$ is non-zero. We define $P_i(x)$ as: $P_i(x) = \Phi \left ( \frac{\left(xW_G \right)_i - kth\_excluding \left(H (x),k, i \right)}{Softplus \left(\left (xW_{noise}\right)_i \right)} \right)$, \\ where $kth\_excluding (H (x), k, i )$ is the $k$th highest component of $H$ excluding the $i$th component. $\Phi$ is the cumulative distribution function of the standard normal distribution.
\item \textbf{Loss Function:} $L = L_{cross\_entropy} + \alpha \cdot (L_{imp} + L_{load})$,
        where $\alpha$ is a hyperparameter.

\end{itemize}            

\section{Experiments}
\subsection{Baselines}
We compare our introduced method with the following research studies (see Section~\ref{related_work_section} for more details of the following studies):

\begin{itemize}
    \item Unimodal Approaches:
    \begin{itemize}
        \item Random Forest \cite{lebedev2014random}
        \item 3D CNNs \cite{payan2015predicting}
    \end{itemize}
    \item Multimodal Approaches
    \begin{itemize}
        \item Stacked Autoencoders \cite{liu2014early}
        \item Multiscale DNN \cite{lu2018mmdnn}
        \item 3D Multiscale CNN \cite{song2021imagefusion} 
        \item 2D \& 3D CNNs \cite{castellano2024automated}
        \item camAD \cite{10822255}
        \item PT DCN \cite{9490307} 
        \item 3D CNN + 3D CLSTM \cite{9098621}
    \end{itemize}
\end{itemize}

\subsection{Experimental Setup}
We perform three classification tasks: AD vs NC, AD vs MCI, and MCI vs NC. To identify the optimal hyperparameters, we utilize the wandb \cite{wandb} framework with random search. The Adam optimizer with an initial learning rate of \(1 \times 10^{-4}\) and a weight decay of 0.1 is used. Also a batch size of 4 is selected, while for regularization, we apply a dropout rate of 0.5. For the number of experts, we set \(n = 5\) and \(k = 4\), while we choose \(a = 0.6\). We split the dataset into a train, validation, and test set (60/20/20\%). Each experiment is trained for 500 epochs. An early stopping strategy is employed with a tolerance of 30 epochs. In this study, the networks are implemented using the PyTorch deep learning framework \cite{paszke2019pytorch}. Experiments are conducted on
a NVIDIA A100 80GB PCIe GPU.

% \begin{table}[ht]
% \centering
% \renewcommand{\arraystretch}{1.4} % Increases row height
% \caption{Optimal hyperparameters used for the experiments.}
% \label{tab:hyperparameters}
% \begin{tabular}{|l|l|}
% \hline
% \textbf{Hyperparameter}    & \textbf{Value}          \\ \hline
% Learning rate (\(\eta\))   & \(1 \times 10^{-4}\)    \\ \hline
% Weight decay               & 0.1                    \\ \hline
% Batch size                 & 4                      \\ \hline
% Dropout rate               & 0.5                    \\ \hline
% Number of experts (\(n\))  & 5                      \\ \hline
% Selected experts (\(k\))   & 4                      \\ \hline
% Parameter \(a\) in loss    & 0.6                    \\ \hline
% Optimizer                  & Adam                   \\ \hline
% \end{tabular}
% \end{table}

\subsection{Evaluation Metrics}

Accuracy (Acc.), sensitivity (Sens.), specificity (Spec.), and AUROC are used as evaluation metrics. The results are reported as the mean $\pm$ standard deviation (SD) across ten runs.

\section{Results}
\subsection{Comparison of our fusion strategies}
Table~\ref{tab:performance} reports the performance of our proposed approach across three binary classification tasks: NC vs MCI, MCI vs AD, and NC vs AD. Advanced fusion methods improve sensitivity in most cases compared to the concatenation mechanism. In comparison with concatenation, GMU gains +4.28\% on NC-MCI (79.71\% vs 75.43\%) and +7.06\% on NC-AD (94.31\% vs. 87.25\%), but is lower on MCI-AD (-2.01\%; 77.23\% vs 79.24\%). Gated self-attention improves sensitivity across all three tasks: +2.92\% on NC-MCI (78.35\%), +2.19\% on MCI-AD (81.43\%), and +5.03\% on NC-AD (92.28\%). Comparing the performance between GMU and gated self-attention, GMU leads on NC-MCI (+1.36\%) and NC-AD (+2.03\%), while gated self-attention clearly wins on MCI-AD (+4.20\%).

We attribute this pattern to GMU’s gating, which emphasizes the most informative modality and suppresses weaker ones (advantageous when one modality dominates and classes are well separated). Gated self-attention, by modeling inter- and intra-modal interactions, better aggregates subtle, complementary cues for the harder MCI-AD discrimination. In contrast, simple concatenation treats every modality equally, so noise from less informative inputs reduces sensitivity.

\begin{table*}[ht]
\scriptsize
\centering
\renewcommand{\arraystretch}{1.2} 
\caption{Performance of our proposed approach. Results are averaged across 10 runs.}
\label{tab:performance}
\begin{tabular}{>{\centering\arraybackslash}p{4cm}|>{\centering\arraybackslash}p{3cm}|>{\centering\arraybackslash}p{2cm}|>{\centering\arraybackslash}p{2cm}|>{\centering\arraybackslash}p{2cm}|>{\centering\arraybackslash}p{2cm}}
\hline
\rowcolor{gray!30}
\textbf{Fusion Method} & \textbf{Task} & \textbf{Acc.} & \textbf{Sens.} & \textbf{Spec.} & \textbf{AUROC} \\ 
\hline
\multirow{3}{*}{Concatenation} & NC vs MCI & $78.25 \pm $3.20 & $75.43 \pm $4.10 & $79.32 \pm $2.10 & $76.56 \pm $3.90 \\
 & MCI vs AD & $80.13 \pm $5.30 &  $79.24 \pm $5.80 &  $81.21 \pm $5.50 &  $76.83 \pm $8.10 \\ 
 & NC vs AD &  $89.52 \pm $3.40 & $87.25 \pm $3.20 & $89.98 \pm $4.1 & $89.64 \pm $2.30 \\ 
\hline
\multirow{3}{*}{GMU} & NC vs MCI & $\mathbf{80.46} \pm 3.90$ &  $\mathbf{79.71} \pm 4.00$ &  $\mathbf{81.76} \pm 3.90$ &  $\mathbf{80.51} \pm 3.50$ \\   
& MCI vs AD & $79.13 \pm $1.10 & $77.23 \pm $3.30 & $81.36 \pm $4.20 & $79.94 \pm $1.50  \\ 
& NC vs AD & $\mathbf{95.47} \pm 2.10$ & $\mathbf{94.31} \pm 3.20$ & $\mathbf{96.73} \pm 1.80$ & $\mathbf{95.41} \pm 2.60$ \\ 
\hline
\multirow{3}{*}{Attention} & NC vs MCI & $80.15 \pm $2.20 & $78.35 \pm $5.40 & $83.56 \pm $2.60 & $77.46 \pm $1.90 \\ 
 & MCI vs AD & $\mathbf{82.08} \pm 2.10$ & $\mathbf{81.43} \pm 1.80$ & $\mathbf{85.24} \pm 2.70$ & $\mathbf{80.48} \pm 3.00$ \\
 & NC vs AD & $91.53 \pm $4.70 & $92.28 \pm $4.40 & $91.07 \pm $4.70 & $92.29 \pm $5.20\\ 
\hline
\end{tabular}
\end{table*}

\subsection{Comparison of our approach with prior work}

Tables~\ref{tab:comp-nc-ad}-\ref{tab:comp-mci-ad} present a comparative analysis of our approach against state-of-the-art techniques.

\textbf{NC vs AD (Table~\ref{tab:comp-nc-ad}).} GMU attains the top Accuracy (95.47\%) and Specificity (96.73\%) among all listed methods, while being second-best in Sensitivity (94.31\%) and AUROC (95.41\%). Relative to the best previously reported numbers per metric, GMU improves Accuracy by 0.08\% (vs. 3D CNNs) and Specificity by 0.07\% (vs. 2D\&3D CNNs); it is slightly below the best Sensitivity by 0.69\% (vs. camAD) and below the best AUROC by 0.99\% (vs. PT DCN).
Across the set of multimodal baselines, GMU improves Accuracy by 0.47-10.87\% and Specificity by 0.07-9.51\%. In Sensitivity, GMU surpasses all but camAD, exceeding the rest by 0.98-14.11\%. For AUROC, GMU is competitive, outperforming 2D\&3D CNNs by +2.41\% and camAD by 0.51\%, while trailing PT DCN by 0.99\%.

\textbf{NC vs MCI.} GMU (ours) achieves 80.46\% in Accuracy, 79.71\% in Sensitivity, 81.76\% in Specificity, and 80.51\% in AUROC (Table~\ref{tab:comp-nc-mci}). Compared to multimodal baselines, GMU improves over Stacked Auto-Encoders by 3.54\% in Accuracy, 5.42\% Sensitivity, and 3.63\% Specificity; relative to 3D CNN + 3D CLSTM, it yields improvements of 1.45\% in Accuracy and 6.27\% in Specificity; against 3D Multi-Scale CNNs, it trails by 4.54\% in Accuracy, 4.98\% in Sensitivity, and 3.84\% in Specificity. Compared with the strongest unimodal baseline, 3D CNNs (92.11\% Accuracy), GMU underperforms by 11.65\% in Accuracy; other metrics were not reported for that method. Overall, GMU surpasses two of the three multimodal baselines in Accuracy and Specificity.

\textbf{MCI vs AD (Table~\ref{tab:comp-mci-ad}).} Our gated self-attention model outperforms the 3D Multi-Scale CNN by 1.28\% in Accuracy and 10.24\% in Sensitivity, with comparable Specificity (-0.70\%), and additionally reports 80.48\% AUROC (not provided by \cite{song2021imagefusion}).

\begin{table}[t]
\scriptsize
\centering
\renewcommand{\arraystretch}{1.3}
\caption{Comparison with prior work on \textbf{NC vs. AD}.}
\label{tab:comp-nc-ad}
\begin{tabular}{lcccc}
\toprule
\textbf{Architecture} & \textbf{Acc.} & \textbf{Sens.} & \textbf{Spec.} & \textbf{AUROC} \\
\midrule
\rowcolor{gray!20}\multicolumn{5}{l}{\textbf{Unimodal approaches (MRI)}} \\
Random Forest \cite{lebedev2014random} & $-$ & $88.60$ & $92.00$ & $-$ \\
3D CNNs \cite{payan2015predicting}     & $95.39$ & $-$ & $-$ & $-$ \\
\midrule
\rowcolor{gray!20}\multicolumn{5}{l}{\textbf{Multimodal approaches}} \\
Stacked AutoEncoders \cite{liu2014early}    & $87.76$ & $88.57$ & $87.22$ & $-$ \\
Multiscale DNN \cite{lu2018mmdnn}            & $84.60$ & $80.20$ & $91.80$ & $-$ \\
3D Multiscale CNN \cite{song2021imagefusion}           & $94.11$ & $93.33$ & $94.27$ & $-$ \\
2D\&3D CNNs \cite{castellano2024automated}   & $95.00$ & $93.33$ & $96.66$ & $93.00$ \\
camAD \cite{10822255}                        & $94.50$ & \textbf{95.00} & $94.90$ & $94.90$ \\
PT DCN \cite{9490307}                        & $92.70$ & $91.70$ & $93.50$ & $\textbf{96.40}$ \\
\midrule
\rowcolor{gray!20}\multicolumn{5}{l}{\textbf{Our best-performing model}} \\
GMU (ours)                                   & \textbf{95.47} & $94.31$ & \textbf{96.73} & 95.41 \\
\bottomrule
\end{tabular}
\end{table}

% ===================== NC vs MCI =====================
\begin{table}[t]
\scriptsize
\centering
\renewcommand{\arraystretch}{1.3}
\caption{Comparison with prior work on \textbf{NC vs. MCI}.}
\label{tab:comp-nc-mci}
\begin{tabular}{lcccc}
\toprule
\textbf{Architecture} & \textbf{Acc.} & \textbf{Sens.} & \textbf{Spec.} & \textbf{AUROC} \\
\midrule
\rowcolor{gray!20}\multicolumn{5}{l}{\textbf{Unimodal approaches (MRI)}} \\
3D CNNs \cite{payan2015predicting}          & \textbf{92.11} & $-$    & $-$    & $-$ \\
\midrule
\rowcolor{gray!20}\multicolumn{5}{l}{\textbf{Multimodal approaches}} \\
Stacked Auto-Encoders \cite{liu2014early}    & $76.92$ & $74.29$ & $78.13$ & $-$ \\
3D Multi-Scale CNNs \cite{song2021imagefusion}           & 85.00 & \textbf{84.69} & \textbf{85.60} & $-$ \\
3D CNN + 3D CLSTM \cite{9098621}             & $79.01$ & $82.35$ & $75.49$ & $-$ \\
\midrule
\rowcolor{gray!20}\multicolumn{5}{l}{\textbf{Our best-performing model}} \\
GMU (ours)                                   & $80.46$ & $79.71$ & $81.76$ & \textbf{80.51} \\
\bottomrule
\end{tabular}
\end{table}

% ===================== MCI vs AD =====================
\begin{table}[t]
\scriptsize
\centering
\renewcommand{\arraystretch}{1.3}
\caption{Comparison with prior work on \textbf{MCI vs. AD}.}
\label{tab:comp-mci-ad}
\begin{tabular}{lcccc}
\toprule
\textbf{Architecture} & \textbf{Acc.} & \textbf{Sens.} & \textbf{Spec.} & \textbf{AUROC} \\
\midrule
\rowcolor{gray!20}\multicolumn{5}{l}{\textbf{Unimodal approaches (MRI)}} \\
3D CNNs \cite{payan2015predicting}          & \textbf{86.84} & $-$    & $-$     & $-$ \\
\midrule
\rowcolor{gray!20}\multicolumn{5}{l}{\textbf{Multimodal approaches}} \\
3D Multi-Scale CNN \cite{song2021imagefusion}           & 80.80 & 71.19 & \textbf{85.94} & $-$ \\
\midrule
\rowcolor{gray!20}\multicolumn{5}{l}{\textbf{Our best-performing model}} \\
Gated self-attention (ours)                   & $82.08$ & \textbf{81.43} & $85.24$ & \textbf{80.48} \\
\bottomrule
\end{tabular}
\end{table}

\subsection{Ablation Study}
In this section, we perform a series of ablation experiments to explore the effectiveness of the introduced approach.

\subsubsection{MoE}
To assess the contribution of the MoE model, we replace it with a simple architecture consisting of three fully connected layers. Results are reported in Table~\ref{without_MoE_Ablation}. Results show that Accuracy drops across all tasks in terms of all fusion strategies (see Table~\ref{tab:performance}).

\begin{table*}[ht]
\scriptsize
\centering
\renewcommand{\arraystretch}{1.3} 
\caption{Ablation Study. Performance without the MoE framework. Results are averaged across 10 runs.}
\label{without_MoE_Ablation}
\begin{tabular}{>{\centering\arraybackslash}p{4cm}|>{\centering\arraybackslash}p{3cm}|>{\centering\arraybackslash}p{2cm}|>{\centering\arraybackslash}p{2cm}|>{\centering\arraybackslash}p{2cm}|>{\centering\arraybackslash}p{2cm}}
\hline
\rowcolor{gray!30}
\textbf{Fusion Method} & \textbf{Task} & \textbf{Acc.} & \textbf{Sens.} & \textbf{Spec.} & \textbf{AUROC} \\ 
\hline
\multirow{3}{*}{Concatenation} & NC vs MCI & $67.40 \pm $0.80 & $76.20 \pm $6.00 & $57.40 \pm $4.20 & $63.10 \pm $4.90 \\ 
 & MCI vs AD & $68.50 \pm $2.80 & $75.20 \pm $5.00 & $59.30 \pm $3.10 & $67.40 \pm $3.90 \\ 
 & NC vs AD & $86.48 \pm $5.20 & $84.32 \pm $4.20 & $87.18 \pm $5.60 & $84.80 \pm $4.80 \\ 
\hline
\multirow{3}{*}{GMU} & NC vs MCI & 70.50 $\pm $2.90 & $74.20 \pm $4.10 & $65.60 \pm $4.80 & $72.30 \pm $3.90 \\ 
 & MCI vs AD & $69.58 \pm $2.70 & $69.88 \pm $3.40 & $67.13 \pm $2.10 & $67.55 \pm $5.60 \\ 
 & NC vs AD & $85.65 \pm $7.20 & $84.51 \pm $5.60 & $88.61 \pm $5.20 & $81.25 \pm $7.80 \\ 
\hline
\multirow{3}{*}{Attention} & NC vs MCI & $68.40 \pm $3.80 & $70.20 \pm $5.10 & $66.80 \pm $4.50 & $67.10 \pm $4.10 \\ 
 & MCI vs AD & $73.45 \pm $2.80 & $75.50 \pm $4.00 & $72.30 \pm $3.50 & $74.30 \pm $2.90 \\ 
 & NC vs AD & $85.00 \pm $8.80 & $84.91 \pm $6.60 & $84.78 \pm $7.20 & $83.10 \pm $7.40 \\ 
\hline
\end{tabular}
\end{table*}

\subsubsection{Unimodal models}
Table \ref{performance} presents the performance metrics of unimodal models (MRI and PET) for the three classification tasks: NC vs MCI, MCI vs AD, and NC vs AD. We observe that unimodal models yield worse results compared to multimodal ones (see Table~\ref{tab:performance}); PET outperforms MRI in all classification tasks.

\begin{table*}[ht]
\scriptsize
\centering
\renewcommand{\arraystretch}{1.5} 
\caption{Ablation Study. Performance of the unimodal models. Results are averaged across 10 runs.}
\label{performance}
\begin{tabular}{>{\centering\arraybackslash}p{4cm}|>{\centering\arraybackslash}p{3cm}|>{\centering\arraybackslash}p{2cm}|>{\centering\arraybackslash}p{2cm}|>{\centering\arraybackslash}p{2cm}|>{\centering\arraybackslash}p{2cm}}
\hline
\rowcolor{gray!30}
\textbf{Fusion Method} & \textbf{Task} & \textbf{Acc.} & \textbf{Sens.} & \textbf{Spec.} & \textbf{AUROC} \\ 
\hline
\multirow{3}{*}{MRI} & NC vs MCI & $67.38 \pm $0.70 & $86.17 \pm $6.00 & $40.35 \pm $8.40 & $63.08 \pm $5.00 \\ 
 & MCI vs AD & $64.29 \pm $4.60 & $67.56 \pm $5.60 &  $62.19 \pm $4.80 &  $61.14 \pm $5.50   \\ 
 & NC vs AD & $75.32 \pm $3.50 & $80.17 \pm $11.10 & $65.31 \pm $10.80 & $76.13 \pm $5.30 \\ 
\hline
\multirow{3}{*}{PET} & NC vs MCI & $72.39 \pm $3.80 & $70.15 \pm $4.80 & $76.21 \pm $4.20 & $73.65 \pm $4.90 \\ 
 & MCI vs AD & $70.81 \pm $2.40 & $68.57 \pm $4.20 & $75.42 \pm $4.10 & $69.60 \pm $5.80 \\ 
 & NC vs AD & $81.10 \pm $1.60 & $81.00 \pm $1.60 & $81.88 \pm $2.60 & $84.00 \pm $6.00   \\ 
\hline
\end{tabular}
\end{table*}

\section{Explainable AI}
\begin{figure}[H]
    \centering
    \includegraphics[width=1\columnwidth]{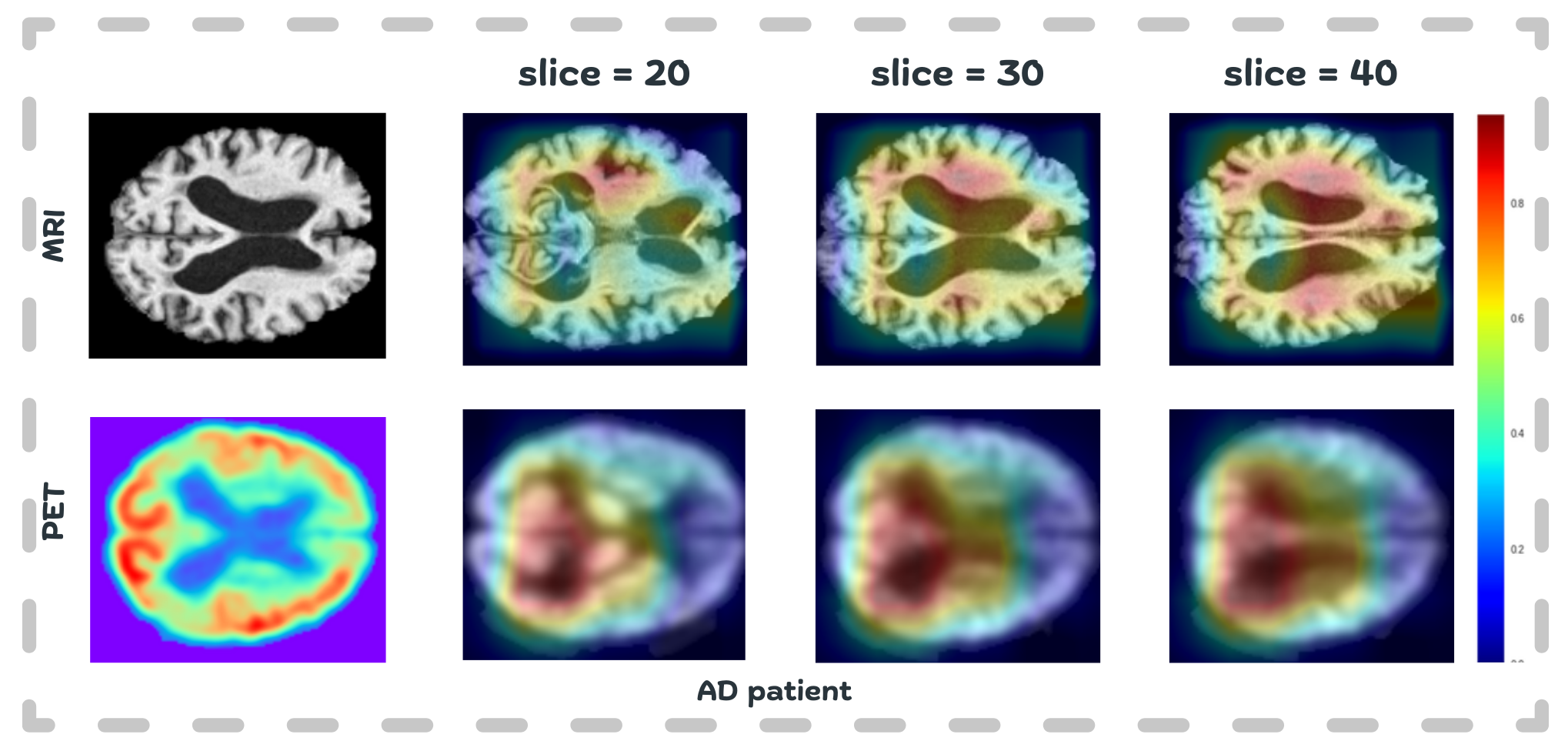} 
    \caption{Grad-CAM results for an AD patient.} 
    \label{fig:gradcam} 
\end{figure}

\noindent Figure~\ref{fig:gradcam} illustrates Grad-CAM visualizations applied to MRI and PET scans of an AD patient. The first column presents the original MRI (top) and PET (bottom) scans, while the subsequent columns display Grad-CAM heatmaps overlaid on different axial slices (\textit{slice = 20, 30, 40}). This visualization aids in interpreting the model's decision-making process by identifying key regions contributing to AD classification. The red regions represent areas that the model considers relevant to AD, with darker shades of red signifying higher importance in the classification decision. We observe that the highlighted regions differ between MRI and PET scans, indicating their complementary nature.

\section{Conclusion}

In this paper, we present the first study utilizing both MRI and PET scans, multimodal fusion methods, and MoE model in a single neural network for the diagnosis of AD and MCI cases. We perform three classification experiments, with experiments demonstrating multiple benefits over state-of-the-art. A series of ablation experiments shows the effectiveness of the introduced approach. Finally, Grad-CAM is used as an explainability method and indicates brain regions contributing to a specific diagnosis.

\textbf{Limitations:}
Our approach depends on labelled datasets. However, obtaining labelled datasets in the healthcare domain is a challenging task. Therefore, more advanced methods are needed to be proposed to tackle the issue of labelled data. Additionally, our approach uses only neuroimaging data, excluding genetic and clinical data.

\textbf{Future Work:}
GANs could be utilized to address the limited dataset size. GANs can be employed for data augmentation, generating synthetic but realistic MRI and PET scans to increase the training set diversity. This could improve the model’s ability to generalize. Self-supervised learning can also be explored to address the issue of limited data. Additionally, neural architecture search and federated learning are two of our future plans.

\section*{Acknowledgments}

The work presented was supported in part by the H2020 European Commission project MES-CoBraD (Grant Agreement No. 965422).

\bibliographystyle{IEEEtran}
\bibliography{references}

\end{document}